\documentclass[11pt]{article}

\usepackage[preprint]{acl}

\usepackage{times}
\usepackage{latexsym}

\usepackage[T1]{fontenc}
\usepackage[utf8]{inputenc}

\usepackage{microtype}

\usepackage{inconsolata}

\usepackage{graphicx}
\usepackage{amsmath}
\usepackage{multirow}
\usepackage{amssymb}
\usepackage{xcolor}
\definecolor{darkgreen}{rgb}{0.0, 0.5, 0.0}
\usepackage{colortbl}
\usepackage{enumitem}

\usepackage{algorithm}
\usepackage{algorithmic}
\usepackage{booktabs}       
\usepackage{multirow}       
\usepackage{geometry}       
\usepackage[table]{xcolor}  
\usepackage{arydshln}

\definecolor{bgOverall}{RGB}{225, 245, 250} 
\definecolor{bgClass}{RGB}{240, 235, 250}   
\definecolor{bgRet}{RGB}{250, 235, 235}   

\title{TabEmbed: Benchmarking and Learning Generalist Embeddings for Tabular Understanding}

\author{
  \textbf{Minjie Qiang\textsuperscript{1,2}\thanks{\ \ Work done at Ant Group.}},
  \textbf{Mingming Zhang\textsuperscript{2}},
  \textbf{Xiaoyi Bao\textsuperscript{3}},
  \textbf{Xing Fu\textsuperscript{2}},
\\
  \textbf{Yu Cheng\textsuperscript{2}},
  \textbf{Weiqiang Wang\textsuperscript{2}},
  \textbf{Zhongqing Wang\textsuperscript{1}},
  \textbf{Ningtao Wang\textsuperscript{2}\thanks{\ \ Corresponding author.}}
\\
\\
  \textsuperscript{1}Natural Language Processing Lab, Soochow University, Suzhou, China\\
  \textsuperscript{2}Ant Group, Hangzhou, China\\
  \textsuperscript{3}The Hong Kong Polytechnic University, Hong Kong, China
\\
  \small{
    \texttt{mjqiang@stu.suda.edu.cn}, \texttt{xiaoyi.bao@connect.polyu.hk}, \texttt{wangzq@suda.edu.cn}
  }\\
  \small{
    \texttt{\{mia.zmm, zicai.fx, cy122623, weiqiang.wwq, ningtao.nt\}@antgroup.com}
  }
}

\begin{document}
\maketitle
\begin{abstract}
Foundation models have established unified representations for natural language processing, yet this paradigm remains largely unexplored for tabular data.
Existing methods face fundamental limitations: LLM-based approaches lack retrieval-compatible vector outputs, whereas text embedding models often fail to capture tabular structure and numerical semantics.
To bridge this gap, we first introduce the Tabular Embedding Benchmark (TabBench), a comprehensive suite designed to evaluate the tabular understanding capability of embedding models.
We then propose TabEmbed, the first generalist embedding model that unifies tabular classification and retrieval within a shared embedding space.
By reformulating diverse tabular tasks as semantic matching problems, TabEmbed leverages large-scale contrastive learning with positive-aware hard negative mining to discern fine-grained structural and numerical nuances.
Experimental results on TabBench demonstrate that TabEmbed significantly outperforms state-of-the-art text embedding models, establishing a new baseline for universal tabular representation learning.
Code and datasets are publicly available at \url{https://github.com/qiangminjie27/TabEmbed} and \url{https://huggingface.co/datasets/qiangminjie27/TabBench}.
\end{abstract}

\section{Introduction}
Recently, foundation models have achieved remarkable success in establishing universal representations for Natural Language Processing~\cite{wang2024improving,qwen3}, such as Retrieval-Augmented Generation (RAG)~\cite{qiang2025exploring}, where dense text embeddings enable efficient semantic search through vector similarity computation. 
However, this unified representation paradigm has not been effectively adapted to tabular data.
Existing research~\cite{ye2025closer,qu2025tabicl,mueller2025mothernet} typically treats tabular classification and retrieval as distinct problems requiring specialized models.
Consequently, the tabular domain lacks a shared embedding space capable of simultaneously addressing all tabular understanding tasks without task-specific architectures.

\begin{figure}[!tp]
	\centering
	\includegraphics[width=\columnwidth]{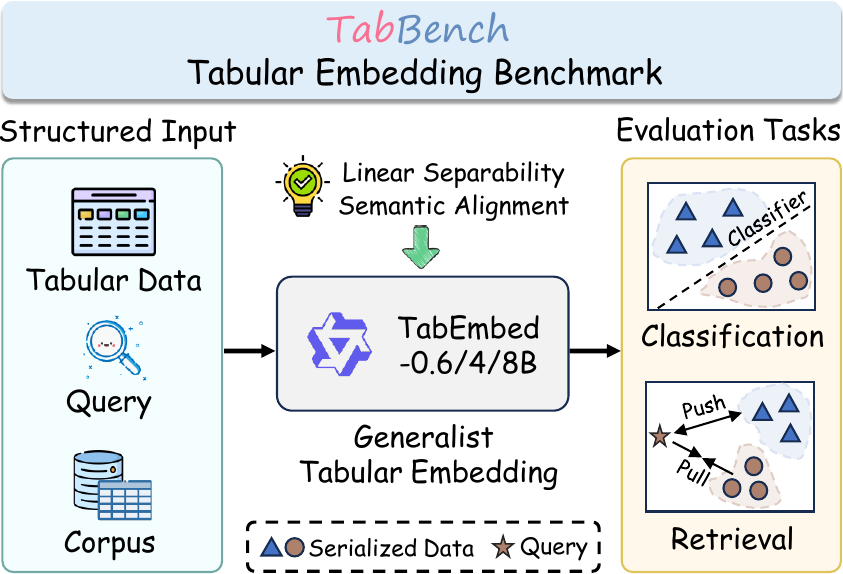}
    	\caption{\label{figure:overview} Overview of \textbf{TabBench} and \textbf{TabEmbed}.}
\end{figure}

Traditional tree-based models excel at tabular classification tasks but are constrained by fixed schemas, rendering them incompatible with zero-shot transfer and retrieval scenarios. 
Recent advances in large language models have shown considerable promise for tabular tasks~\cite{gardner2024large,ye2025closer,qu2025tabicl}.
However, these methods do not produce the dense, fixed-dimensional vectors required for vector databases and downstream retrieval applications.
While general-purpose text embedding models~\cite{zhang2025jaspertokencompression600mtechnicalreport,yu2025qzhouembeddingtechnicalreport,qwen3embedding} can generate such embeddings with remarkable success in text domains, they treat serialized tables as unstructured text, often failing to capture essential structural logic such as numerical magnitude and column-specific semantics.
These constraints motivate the development of a generalist tabular embedding model that inherently understands tabular structure to handle various tabular understanding tasks within a shared embedding space.

However, training such a tabular embedding model presents three significant challenges.
First, the absence of benchmarks specifically designed for tabular embeddings hinders systematic evaluation.
Second, existing contrastive learning paradigms in the tabular domain are inadequate for unified understanding.
Prior works typically rely on a row-to-row contrastive objective, where a data row serves as the anchor and is aligned with augmented views or other rows of the same class (e.g., SCARF~\cite{bahri2021scarf}).
While this paradigm effectively separates classes, it forces the embedding space to collapse into coarse class clusters.
By indiscriminately pulling together rows with divergent feature values simply because they share a target label, the model discards fine-grained structural semantics, logical constraints, and numerical magnitudes.
Consequently, these representations fail to support precise semantic matching and retrieval. 
Finally, unifying classification and retrieval within a shared embedding space is non-trivial.
Retrieval relies on semantic ranking to identify relevant data, whereas classification requires precise decision boundaries for label prediction.

The core value proposition of TabEmbed is to provide a universal, schema-agnostic representation that unifies diverse tabular tasks into a shared semantic space.
This is an objective that traditional schema-bound models (e.g., XGBoost) cannot achieve without task-specific retraining.
As shown in Figure~\ref{figure:overview}, we first introduce TabBench, a comprehensive evaluation suite assessing numerical reasoning and retrieval capabilities.
Then we propose TabEmbed, an embedding model that unifies classification and retrieval within a shared embedding space. 
To train this model, we depart from the suboptimal row-to-row paradigm and introduce a unified language-to-row contrastive framework.
By synthesizing task-adaptive natural language queries as anchors, we reformulate diverse tasks into semantic matching problems.
Enhanced by positive-aware hard negative mining, TabEmbed is compelled to discern fine-grained schema differences.
Extensive experiments on TabBench demonstrate that TabEmbed significantly outperforms state-of-the-art text embeddings, establishing a new baseline for tabular understanding.

\section{The Tabular Embedding Benchmark}
\label{section:benchmark}
\begin{figure}[!tp]
	\centering
	\includegraphics[width=\columnwidth]{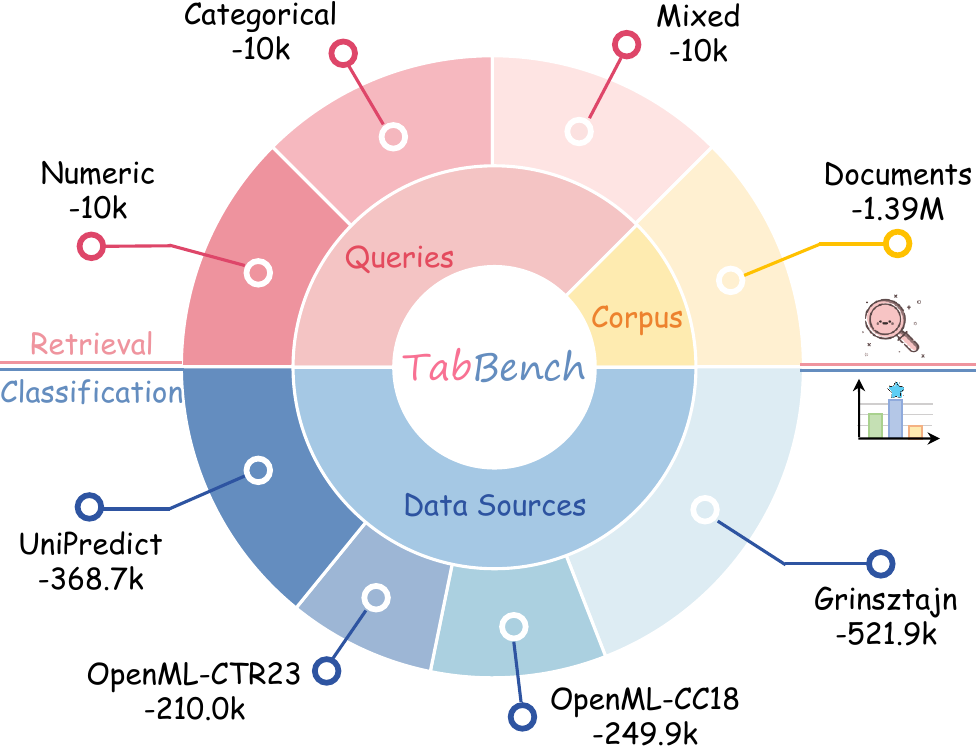}
	\caption{\label{figure:tabbench_composition} Data composition and statistics of TabBench.}
\end{figure}

To rigorously evaluate the capabilities of embedding models in tabular understanding, we introduce the \textbf{Tabular Embedding Benchmark (TabBench)}.
Building upon the high-quality data curation of the \texttt{tabula-8b-eval-suite}~\cite{gardner2024large}, TabBench provides a comprehensive framework to assess two critical dimensions of tabular representation:
linear separability (via classification) and semantic alignment (via retrieval).
The benchmark aggregates diverse datasets from four authoritative repositories: 
\textbf{Grinsztajn}~\cite{grinsztajn2022tree}, \textbf{OpenML-CC18}~\cite{bischl2017openml}, \textbf{OpenML-CTR23}~\cite{fischer2023openml}, and \textbf{UniPredict}~\cite{wang2023unipredict}.
The detailed composition of TabBench is illustrated in Figure~\ref{figure:tabbench_composition}.
We implement a standardized pipeline for data serialization, task construction, and quality filtering.

\subsection{Data Serialization}
\label{section:Serialization}
Bridging the modality gap between structured tabular data and large language models requires an effective serialization strategy~\cite{hegselmann2023tabllm, gardner2024large}.
Formally, let a tabular row be represented as an ordered sequence of feature-value pairs $\mathbf{x} = ((h_1, v_1), \dots, (h_C, v_C))$, where $h_j$ denotes the column header and $v_j$ is the corresponding cell value, with $C$ being the total number of columns.
We define a serialization function $\mathcal{S}: \mathcal{X} \rightarrow \mathcal{T}$ that maps $\mathbf{x}$ from the tabular space $\mathcal{X}$ to a natural language sequence in the text space $\mathcal{T}$ via string concatenation:
\begin{equation}
    \mathcal{S}(\mathbf{x}) = \bigoplus_{j=1}^{C} \left( \text{``The } h_j \text{ is } \tilde{v}_j\text{.''} \right),
\end{equation}
where $\oplus$ denotes the string concatenation operator, and $\tilde{v}_j$ represents the pre-processed string value.
To maintain token efficiency and align with the context constraints of mainstream embedding models, we filter out rows that surpass the predefined maximum sequence length.
Further details regarding the pre-processing and standardization of heterogeneous tabular data (e.g., numeric, temporal, and binary fields) are provided in Appendix~\ref{appendix:heterogeneous_data}.

\subsection{Evaluation Tasks}
\label{section:tasks}

We formulate two distinct tasks to comprehensively evaluate the versatility of the learned embeddings within a shared vector space $\mathcal{Z} \in \mathbb{R}^d$. Let $f_\theta(\cdot)$ denote the embedding model that maps an input text sequence to a $d$-dimensional dense vector.

\paragraph{Tabular Classification}
This task evaluates the linear separability of the embeddings. We construct the evaluation suite by treating each source dataset as an independent classification task. Specifically, for a given tabular row, the input is the serialized text of its feature columns $\mathcal{S}(\mathbf{x}_i)$, and the output to predict is its corresponding discrete target label $y_i \in \mathcal{Y}$. Formally, given a dataset $\mathcal{D} = \{(\mathcal{S}(\mathbf{x}_i), y_i)\}_{i=1}^N$, we extract the frozen representations $\mathbf{z}_i = f_\theta(\mathcal{S}(\mathbf{x}_i))$. We then train an independent Logistic Regression classifier $g_\omega: \mathbb{R}^d \rightarrow \mathcal{Y}$ parameterized by $\omega$ for each dataset on top of the embeddings, optimized via:
\begin{equation}
    \hat{\omega} = \arg\min_{\omega} \frac{1}{|\mathcal{D}_{\text{train}}|} \sum_{i \in \mathcal{D}_{\text{train}}} \mathcal{L}_{\text{CE}}(g_\omega(\mathbf{z}_i), y_i),
\end{equation}
where $\mathcal{L}_{\text{CE}}$ denotes the cross-entropy loss. To ensure evaluation quality, we apply a strict filtering protocol: datasets are excluded if the label cardinality $|\mathcal{Y}| > 50$ or the label-to-sample ratio $|\mathcal{Y}| / N > 0.1$. For qualified datasets, we employ stratified sampling to partition data into training and testing splits, guaranteeing a minimum of two samples per class to mitigate cold-start issues for rare classes.

\paragraph{Tabular Retrieval}
\label{section:query_generation}
Unlike classification, which assesses intra-dataset separability, the retrieval task evaluates the model's ability to align natural language queries with serialized rows across a heterogeneous global corpus $\mathcal{U}$. We construct $\mathcal{U}$ by aggregating rows from all datasets, capping each dataset's contribution at 10,000 samples to prevent distribution dominance.

To simulate realistic user intent, we propose a \textit{seed-based query generation} pipeline. For a given ``seed row'' in the corpus, we generate a natural language query $q$ following the template: \textit{``Find records where $c_1$ and $\dots$ and $c_k$''}. This corresponds to a logical constraint condition $\Phi_q = c_1 \land c_2 \land \dots \land c_k$, where each $c_j$ represents an attribute constraint. The retrieval system ranks documents $d \in \mathcal{U}$ based on the cosine similarity score:
\begin{equation}
    s(q, d) = \frac{f_\theta(q) \cdot f_\theta(d)}{\|f_\theta(q)\| \|f_\theta(d)\|}.
\end{equation}
Let $\Phi_q(d) \in \{\text{True}, \text{False}\}$ denote whether document $d$ satisfies the logical constraints in $\Phi_q$. The goal is to retrieve the ideal target set $\mathcal{R}_q = \{d \in \mathcal{U} \mid \Phi_q(d) = \text{True}\}$. Based on the type of constraints, we define three query categories of increasing complexity:
\begin{itemize}[leftmargin=.15in]
\item \textbf{Categorical Queries:} Assess exact-match semantics. Each constraint $c_j$ enforces strict equality on discrete features (e.g., \textit{``Status is Active''}).
\item \textbf{Numeric Queries:} Test the understanding of magnitude and ranges. Each constraint $c_j$ is generated by sampling a relational operator $\sim \, \in \{>, <, =\}$ and perturbing the original feature value (e.g., \textit{``Price < 50.25''}).
\item \textbf{Mixed Queries:} Evaluate complex reasoning by combining numeric and categorical constraints derived from the same row (e.g., \textit{``Status is Active $\land$ Price < 50.25''}).
\end{itemize}

To ensure benchmark validity, we perform symbolic verification for every generated query, retaining only valid queries where the target set cardinality satisfies $|\mathcal{R}_q| \ge 5$. This process yields a balanced evaluation set, where each query contains 1 to 3 conditions (i.e., $k \in \{1, 2, 3\}$), covering diverse logical complexities.

\begin{figure*}[!tp]
	\centering
	\includegraphics[width=\textwidth]{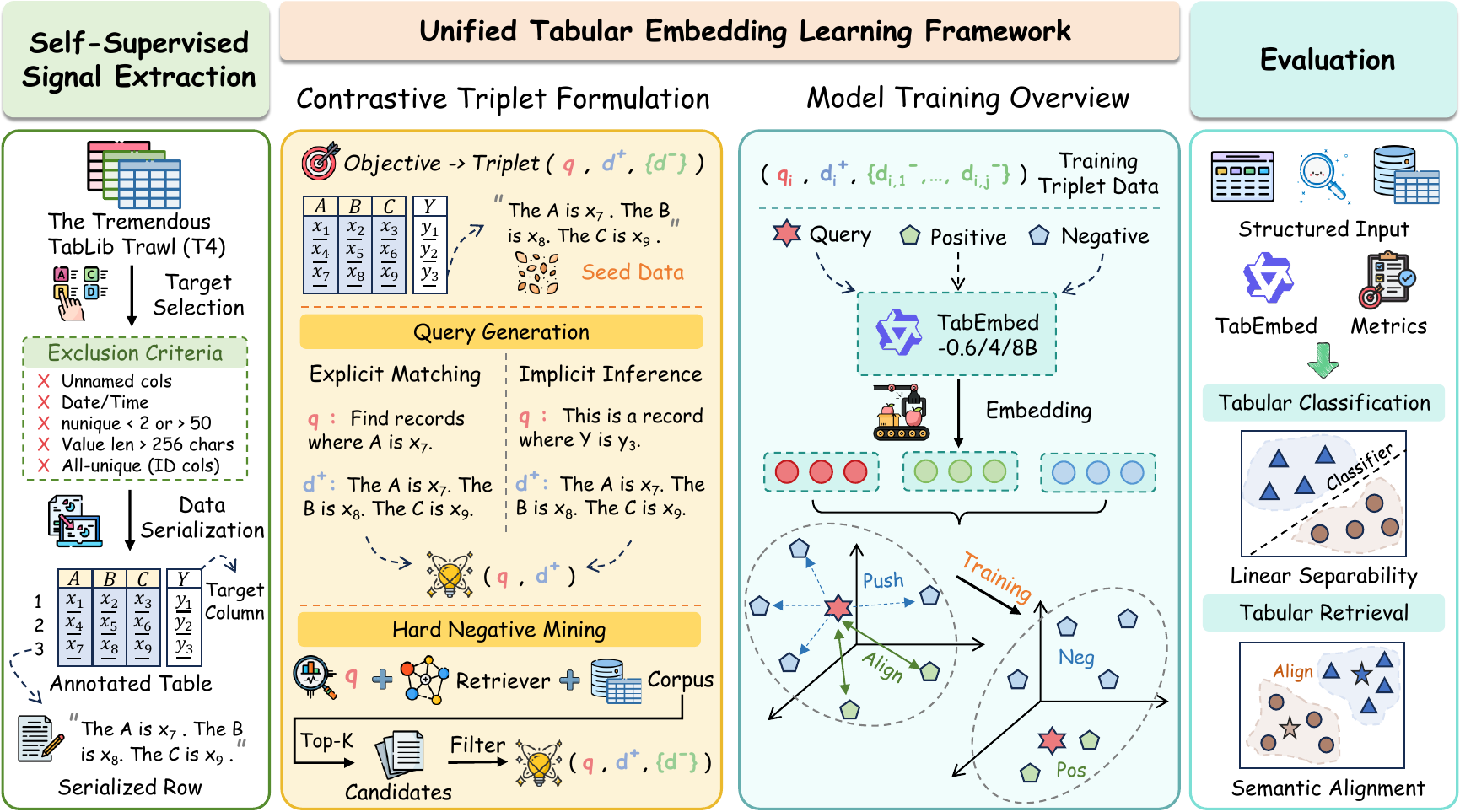}
	\caption{\label{figure:framework} The overall framework of TabEmbed. }
\end{figure*}

\section{TabEmbed: Unified Tabular Embedding Learning}
\label{sec:tabembed_training}

To bridge the gap between structured data and semantic representation, we propose \textbf{TabEmbed}, a generalist embedding model trained within a unified framework that learns tabular representations by casting disparate downstream tasks into a shared contrastive paradigm.
The overall framework is illustrated in Figure~\ref{figure:framework}.
Leveraging the massive scale of the T4 dataset~\cite{gardner2024large}, we propose a novel language-to-row contrastive learning approach.
Unlike conventional tabular methods that rely on row-to-row alignment, which often causes semantic collapse into coarse categories, our framework synthesizes natural language queries as anchors to construct diverse contrastive triplets. 
This strategy unifies disparate downstream capabilities into a shared semantic space while preserving fine-grained tabular structures.

\subsection{Self-Supervised Signal Extraction}
\label{sec:curation}
Since the T4 corpus lacks explicit task annotations, we employ an automated pipeline to transform raw tables into self-supervised training instances. 
We first dynamically identify a target column $y$ within each table to serve as the prediction signal.
To ensure the quality of these self-supervised signals, we apply a rigorous filtering protocol to exclude non-informative attributes (e.g., identifiers, timestamps) and prioritize targets with clear semantic boundaries.
The Detailed pipeline is provided in Appendix~\ref{appendix:target_selection}.
To prevent information leakage and compel the model to learn latent dependencies, we apply a \textit{target-masked serialization} strategy. 
Specifically, we strictly exclude the selected target $y$ from the feature set and apply the serialization function $\mathcal{S}$ (defined in Section~\ref{section:Serialization}) to the remaining columns.
This yields the serialized row $d = \mathcal{S}(\mathbf{x}_{-y})$, ensuring that the embedding captures the row's semantic content without revealing the ground truth label. 

\begin{table*}[!tp]
\centering

\scalebox{0.96}{
\begin{tabular}{
    l 
    c 
    >{\columncolor{bgOverall}}c 
    >{\columncolor{bgClass}}c 
    >{\columncolor{bgClass}}c 
    >{\columncolor{bgRet}}c 
    >{\columncolor{bgRet}}c
}
\toprule

\rowcolor{white}
 & 
 & 
 & 
\multicolumn{2}{c}{\textbf{Classification}} & 
\multicolumn{2}{c}{\textbf{Retrieval}} \\

\cmidrule(lr){4-5} \cmidrule(lr){6-7}

\rowcolor{white}
\multirow{-2}{*}[0.35ex]{\textbf{Embedding Models}} & 
\multirow{-2}{*}[0.35ex]{\textbf{\#Params}} & 
\multirow{-2}{*}[0.35ex]{\textbf{Overall}} & 
\textbf{Accuracy} & \textbf{F1} & \textbf{MRR@10} & \textbf{nDCG@10} \\
\midrule

Jina-Embeddings-v3          & 0.6B & 41.48 & 60.33 & 46.11 & 32.49 & 26.98 \\
Jasper-Token-Compression    & 0.6B & 42.75 & 61.25 & 47.69 & 33.56 & 28.50 \\
Qwen3-Embedding-0.6B        & 0.6B & 44.92 & 62.81 & 50.32 & 36.00 & 30.56 \\
\hdashline
\textbf{TabEmbed-0.6B (Ours)} & 0.6B & \textbf{65.27} & \textbf{67.16} & \textbf{56.56} & \textbf{71.72} & \textbf{65.64} \\

\midrule

F2LLM-4B                    & 4B & 48.02 & 64.92 & 52.48 & 40.60 & 34.08 \\
Octen-Embedding-4B          & 4B & 48.62 & 65.36 & 53.64 & 40.97 & 34.51 \\
Qwen3-Embedding-4B          & 4B & 48.91 & 65.09 & 52.72 & 42.04 & 35.76 \\
\hdashline
\textbf{TabEmbed-4B (Ours)} & 4B & \textbf{70.71} & \textbf{69.51} & \textbf{59.75} & \textbf{79.33} & \textbf{74.25} \\

\midrule

SFR-Embedding-Mistral        & 7B & 49.42 & 64.28 & 50.75 & 44.23 & 38.41 \\
Linq-Embed-Mistral          & 7B & 50.74 & 66.06 & 53.33 & 44.65 & 38.92 \\
GTE-Qwen2-7B-Instruct       & 7B & 51.27 & 64.67 & 51.76 & 47.44 & 41.19 \\
Qwen3-Embedding-8B          & 8B & 48.03 & 65.08 & 52.81 & 40.06 & 34.16 \\
\hdashline
\textbf{TabEmbed-8B (Ours)} & 8B & \textbf{71.62} & \textbf{69.88} & \textbf{60.19} & \textbf{80.58} & \textbf{75.83} \\

\bottomrule
\end{tabular}
}
\caption{\textbf{Tabular Embedding Benchmark (TabBench) Leaderboard.} We evaluate TabEmbed against state-of-the-art generalist text embedding models across three parameter scales. The best results are highlighted in \textbf{bold}.}
\label{tab:main_results}
\end{table*}

\subsection{Contrastive Triplet Formulation}
\label{sec:triplets}
To overcome the limitations of traditional row-to-row instance discrimination, we formulate tabular representation learning as a language-to-row matching problem.
Specifically, we optimize similarity within cross-modal triplets $(q, d^+, \{d^-\})$, where the anchor $q$ is a dynamically generated natural language query expressing a specific tabular constraint or class intent, $d^+$ is the corresponding serialized row satisfying $q$, and $\{d^-\}$ are hard negatives.
We construct these queries to cover both explicit signal matching and implicit semantic inference.

\subsubsection{Task-Adaptive Query Generation}
We generate synthetic queries $q$ to model two complementary tasks using a shared data format:

\paragraph{Tabular Retrieval (Explicit Matching)}
The retrieval task aligns natural language constraints with rows that satisfy them. We leverage the query generation pipeline detailed in Section~\ref{section:query_generation}, which samples subsets of attributes from $\mathbf{x}_{-y}$ to form logical conditions spanning both numerical and categorical fields. 
Formally, for a serialized row $d^+$, we generate a query $q_{\text{ret}}$ describing specific attribute constraints (e.g., \textit{``Find records where Status is Active and Price less than 50.25''}).
This forces the model to align natural language constraints with specific attribute values present in the input.

\paragraph{Tabular Classification (Implicit Inference)}
The classification task aligns abstract label descriptions with rows that imply those labels. Unlike retrieval, the query content (the value of target $y$) is absent from the input $d^+$ and must be inferred solely from the correlations among the remaining features.
For a hidden target column $y$ with value $v$, we construct a descriptive label query $q_{\text{cls}}$ (e.g., \textit{``This is a record where $y$ is $v$.''}).
This formulation encourages the model to cluster rows based on latent predictive features rather than surface-level token overlap.

\subsubsection{Positive-Aware Hard Negative Mining}
% Discriminative embedding learning relies heavily on the quality of negative samples. 
Simple in-batch negatives are insufficient for distinguishing numerically similar values or closely related classes.
We implement an offline Hard Negative Mining strategy using a lightweight dense retriever (\texttt{Qwen3-Embedding-0.6B}).
For every query $q$, we retrieve the Top-$K$ candidates from the global corpus.
Crucially, we employ a Positive-Aware Filtering mechanism: we strictly retain only those candidates that possess high semantic similarity to the query but explicitly violate the retrieval condition or belong to a different class label.
These mined hard negatives $d^-$ constitute the set of samples that are most easily confused with the positive $d^+$, ensuring the model learns sharp decision boundaries.

\subsection{Training Objective}
We optimize our model using the contrastive learning loss. 
Given a batch $\mathcal{B}$ containing $B$ triplets $(q_i, d^+_i, \{d^-_{i,j}\}_{j=1}^H)$, where $H$ is the number of mined hard negatives per query, the objective for query $q_i$ is defined as:

\begin{equation}
    \mathcal{L}_i = -\log \frac{e^{s(q_i, d^+_i)/\tau}}{e^{s(q_i, d^+_i)/\tau} + \sum_{d \in \mathcal{N}_i} e^{s(q_i, d)/\tau}},
\end{equation}

where $s(\cdot, \cdot)$ denotes the cosine similarity (as defined in Section~\ref{section:tasks}), $\mathcal{N}_i$ includes both the $H$ specific hard negatives and the in-batch negatives from other queries in $\mathcal{B}$, and $\tau$ is a temperature hyperparameter.
This unified objective fosters a shared embedding space capable of generalizing across heterogeneous tabular understanding tasks.

\section{Experiments}
\subsection{Implementation Details}
We initialize TabEmbed using the Qwen3-Embedding family~\cite{qwen3embedding} across three scales: 0.6B, 4B, and 8B parameters. This selection allows us to evaluate the scalability of our unified training paradigm across varying computational regimes.
The models are optimized using a contrastive learning objective within the Sentence-Transformers framework.
We conduct evaluations on our proposed TabBench, with dataset statistics detailed in Figure~\ref{figure:tabbench_composition}.
To construct the training data, we curate a balanced mixture of 500,000 retrieval and 100,000 classification contrastive triplets from the T4 dataset.
For evaluation metrics, we report \textbf{Accuracy} and \textbf{F1-Score} for the tabular prediction task, and \textbf{MRR@10} and \textbf{nDCG@10} for the tabular retrieval task.
To provide a holistic measure of generalist capabilities, we also report an \textbf{Overall} score, computed as the macro-average of these four individual metrics.
Further implementation details and evaluation protocols are provided in Appendix~\ref{appendix:implementation_details}.

\subsection{Main Results}
We evaluate TabEmbed on TabBench against a comprehensive suite of ten generalist text embedding models spanning three parameter scales (\textbf{0.6B}, \textbf{4B}, and \textbf{7B-8B}). Detailed specifications and citations for all baseline models are provided in Appendix~\ref{appendix:baselines}.

Table~\ref{tab:main_results} presents the performance evaluation.
The results demonstrate that TabEmbed achieves state-of-the-art performance across all parameter scales, significantly surpassing existing text embedding models.
In \textbf{Tabular Retrieval}, TabEmbed yields substantial improvements, with the 0.6B model surpassing its Qwen3 backbone by over 35 points in MRR@10.
This indicates that our unified contrastive learning paradigm effectively bridges the semantic gap between natural language queries and structured data, addressing a capability largely absent in text embeddings.
In \textbf{Tabular Classification}, TabEmbed consistently improves accuracy and F1 scores, suggesting that the learned representations capture the fine-grained decision boundaries essential for linear separability.
Crucially, our method exhibits remarkable parameter efficiency.
TabEmbed-0.6B outperforms all baselines on the aggregate metric, including those in the 7B and 8B regimes.
This finding suggests that domain-specific contrastive learning is more critical for tabular understanding than model scaling alone.
Nevertheless, scaling TabEmbed from 0.6B to 8B yields consistent performance gains, confirming that our unified paradigm effectively leverages the capacity of larger foundation models to establish a new performance standard for tabular representation.

\subsection{Performance on Diverse Backbones}
To investigate the universality and robustness of our proposed training paradigm, we extend our evaluation beyond the Qwen3 family to a diverse set of backbone architectures.
Specifically, we apply the unified contrastive learning paradigm to eight distinct foundation models, spanning different architectures (e.g., Qwen3, Mistral, and XLM-RoBERT) and parameter scales (ranging from 0.6B to 8B).
We compare the performance of these models before and after applying our training framework, utilizing the original performance as baselines.

\begin{figure}[!tp]
	\centering
	\includegraphics[width=\columnwidth]{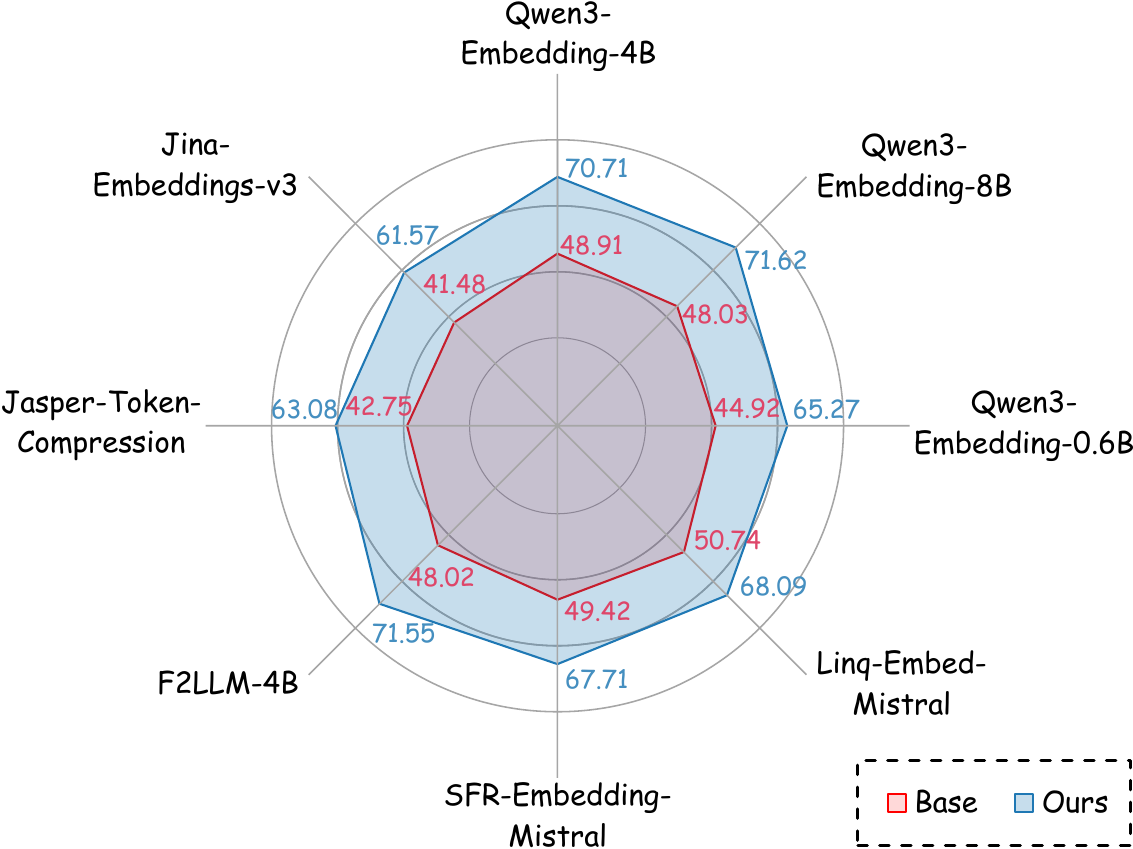}
	\caption{\label{figure:universality} Performance comparison across backbone architectures using our proposed training paradigm. The performance metric is the Overall average score.}
\end{figure}

As illustrated in Figure~\ref{figure:universality}, our approach consistently yields substantial performance improvements across all evaluated backbones, regardless of their architectural design or pre-training objective.
Notably, models based on the Qwen3 architecture (e.g., F2LLM-4B) and the Mistral architecture (e.g., Linq-Embed-Mistral) exhibit significant enhancements, with Qwen3-Embedding-4B achieving the most significant improvement, surging from 48.91 to 70.71.
Even for Jina-Embeddings-v3, which relies on an encoder-only XLM-RoBERT encoder architecture, our method achieves a remarkable gain of over 20 points (rising from 41.48 to 61.57).
These results demonstrate that the improvements stem from the unified contrastive data paradigm rather than model-specific inductive biases, confirming that our paradigm effectively equips diverse text-based foundation models with generalized tabular understanding capabilities.

\section{Analysis and Discussion}

\subsection{Fine-grained Analysis on Retrieval Capabilities}
While the aggregate metrics demonstrate the overall superiority of TabEmbed, it is crucial to understand how the model behaves under different semantic modalities and logical complexities.
To this end, we conduct a fine-grained breakdown of the retrieval performance on the Qwen3-Embedding-0.6B backbone, categorizing the test queries by type (Numeric, Categorical, and Mixed) and the number of logical constraints (from 1 to 3).

\begin{figure}[!tp]
	\centering
	\includegraphics[width=\columnwidth]{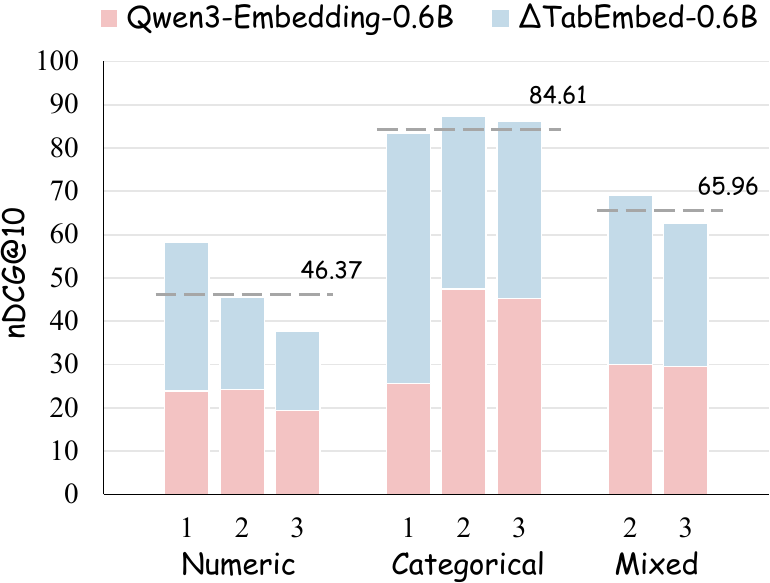}
	\caption{\label{figure:retrieval_breakdown} Fine-grained retrieval performance on TabBench (nDCG@10). The dashed lines indicate the average performance of TabEmbed for each query type.}
\end{figure}

As illustrated in Figure~\ref{figure:retrieval_breakdown}, TabEmbed achieves consistent and substantial improvements across all query scenarios, yet the difficulty varies significantly by task type.
The dashed lines representing the average performance reveal an inherent hierarchy of difficulty: \textit{Categorical} queries are the most solvable (84.61), followed by \textit{Mixed} (65.96), with \textit{Numeric} queries presenting the greatest challenge (46.37).
Crucially, the baseline model exhibits severe limitations in handling numerical queries, often failing to capture magnitude and range relationships.
In contrast, TabEmbed contributes a massive performance gain in the Numeric category, effectively bridging the gap between text-based retrieval and numerical reasoning.
Furthermore, regarding logical complexity, we observe that performance generally correlates with the number of constraints.
For instance, in the Numeric setting, performance naturally decreases as the number of conditions increases from 1 to 3.
Despite this increased difficulty, TabEmbed maintains robust performance, validating its ability to handle complex, multi-condition logical intersections within the embedding space.

\subsection{Numerical Sensitivity Analysis}
\label{sec:numerical_sensitivity}

Standard text embedding models often treat numbers as arbitrary tokens, lacking awareness of magnitude and inequality.
To investigate whether TabEmbed has acquired genuine numerical reasoning capabilities beyond surface-level token matching, we conduct a \textbf{Numerical Sensitivity Test}.
Specifically, for a given query containing a numerical constraint (e.g., $q=$ \textit{``Revenue greater than 500''}), we generate a sequence of candidate values $x$ ranging from small to large.
We then compute the Spearman correlation ($\rho$) between the cosine similarity $\text{sim}(q, x)$ and the ground truth logical satisfaction (i.e., the ideal curve should step up when $x > 500$).

\begin{figure}[!tp]
	\centering
	\includegraphics[width=\columnwidth]{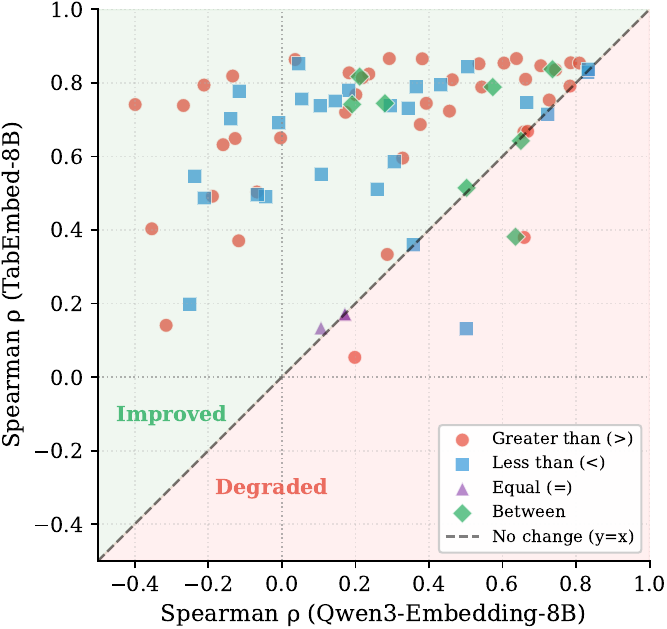}
	\caption{\label{figure:numerical_sensitivity} Pairwise comparison of numerical sensitivity between the baseline and TabEmbed. Each point represents a distinct test case, plotted by the Spearman correlation ($\rho$) between similarity scores and ground truth logic. Points in the green region indicate TabEmbed aligns significantly better with numerical constraints.}
\end{figure}

Figure~\ref{figure:numerical_sensitivity} visualizes the pairwise comparison of these correlation coefficients across diverse test cases, including inequalities ($>, <$), equality ($=$), and range queries (\textit{Between}).
The results reveal a distinct performance gap: the baseline Qwen3-Embedding (X-axis) frequently exhibits near-zero or weakly positive correlations, suggesting it struggles to distinguish between numerically valid and invalid candidates.
In contrast, TabEmbed (Y-axis) shifts the majority of test cases into the upper-left ``Improved'' region, with many cases achieving high correlations ($\rho > 0.8$).
This substantial shift indicates that our model has successfully internalized numerical semantics, mapping mathematically close or logically valid values to closer proximity in the vector space.
Detailed visualizations of similarity curves are provided in Appendix~\ref{appendix:numeric_curves}.

\subsection{Visualization of Embedding Spaces}
To provide a qualitative assessment of the learned representations, we project the high-dimensional embeddings into a 2D space using PCA and t-SNE.
We visualize the geometric structures for both classification and retrieval tasks, comparing the baseline Qwen3-Embedding-8B against TabEmbed-8B.
To quantify the clustering quality, we report the \textit{Cluster Ratio}, defined as the ratio of inter-cluster distance to intra-cluster distance, where a higher ratio indicates better separability.

\begin{figure}[!tp]
	\centering
	\includegraphics[width=\columnwidth]{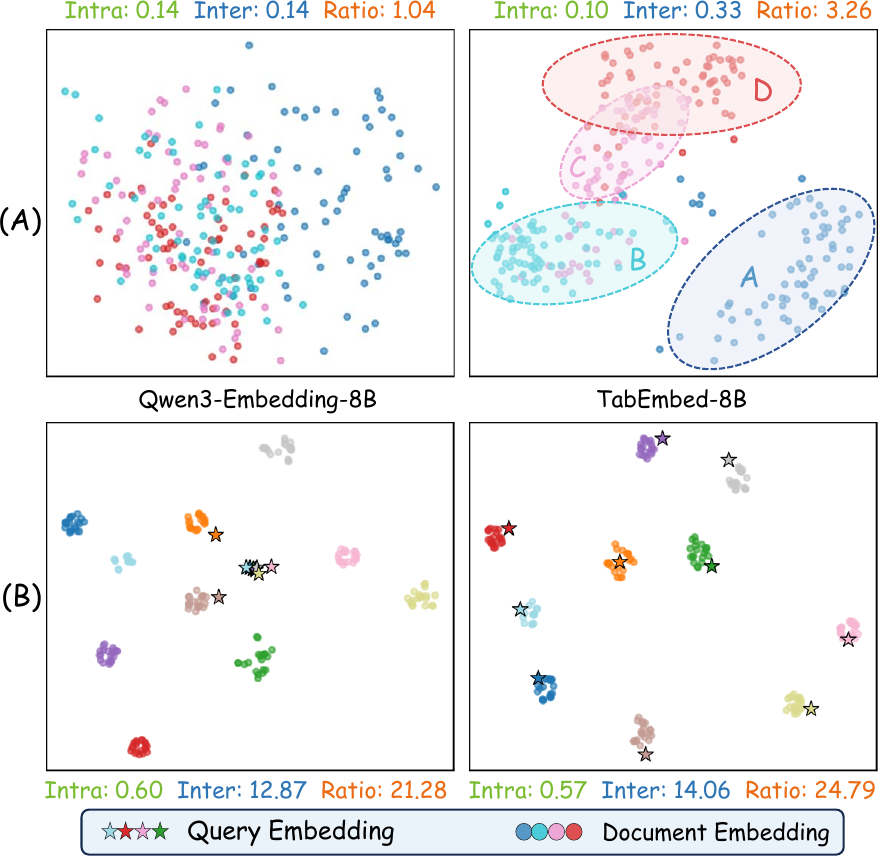}
    \caption{\label{figure:visualization} Visualization comparing Qwen3-Embedding-8B (left) and TabEmbed-8B (right) on tabular classification (A) and tabular retrieval (B) tasks. We report the Cluster Ratio to quantify the clustering quality.}
\end{figure}

Figure~\ref{figure:visualization}(A) illustrates the feature space for classification.
The baseline exhibits a highly entangled distribution (Ratio: 1.04) with significant overlap between classes, suggesting a failure to capture discriminative boundaries.
In contrast, TabEmbed effectively disentangles these classes into well-separated clusters (regions A-D), substantially increasing the Cluster Ratio to 3.26.
This confirms that our contrastive paradigm imparts linear separability to the embedding space, enabling efficient downstream classification.

Figure~\ref{figure:visualization}(B) visualizes semantic alignment for retrieval.
Although the baseline exhibits partial alignment capabilities, many queries ($\star$) remain drifting away from their target document clusters.
In contrast, TabEmbed consistently anchors queries within their corresponding groups and pulls relevant documents tighter around query centers, producing significantly more compact clusters (Intra: 0.60 $\to$ 0.57) and higher separability (Ratio: 21.28 $\to$ 24.79). 
This demonstrates that TabEmbed learns a precise alignment between natural language constraints and structured tabular data, effectively correcting the misalignment observed in the baseline.

\subsection{Robustness to Irrelevant Table Columns}
Real-world tabular data is often characterized by high dimensionality, where a user's query typically targets only a small subset of columns (e.g., filtering by \textit{Price} and \textit{City}) while ignoring numerous irrelevant attributes.
Standard text embedding models are susceptible to \textbf{semantic dilution}, where irrelevant text diminishes the weight of target information in high-dimensional tables.
To evaluate robustness against such structural noise, we incrementally inject up to 30 irrelevant columns into documents initially containing 15 columns and observe the degradation in MRR@10.

\begin{figure}[!tp]
	\centering
	\includegraphics[width=\columnwidth]{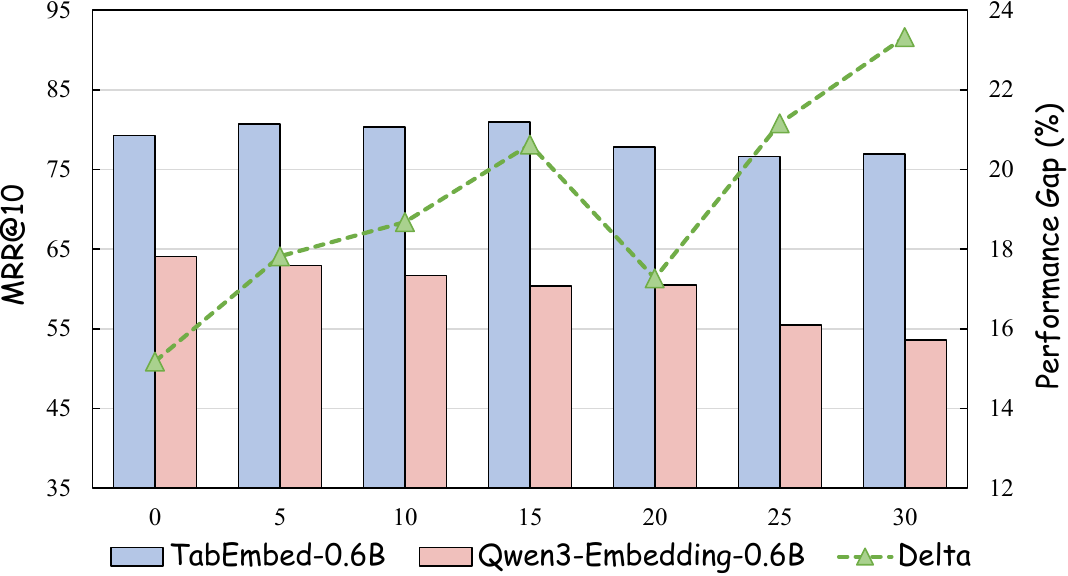}
	\caption{\label{figure:noise_robustness} Robustness analysis against irrelevant table columns. We incrementally inject noise columns (0 to 30) into the documents while maintaining fixed queries.}
\end{figure}

As shown in Figure~\ref{figure:noise_robustness}, the baseline Qwen3-Embedding exhibits a marked sensitivity to noise. Its performance declines steadily as the number of irrelevant columns increases, dropping from $\sim$64\% to below 55\%.
This confirms that without structural awareness, the model struggles to attend to the relevant signal amidst a growing volume of noise tokens.
In contrast, TabEmbed demonstrates exceptional stability, consistently maintaining an MRR@10 above 75\% even when 30 irrelevant columns are added.
Crucially, the green dashed line highlights that the performance gap ($\Delta$) between the two models widens monotonically from $\sim$15\% at the noise-free baseline to over 23\% at the maximum noise level.
This result suggests that TabEmbed has effectively learned an implicit structural attention mechanism, enabling it to selectively align query constraints with matching columns while filtering out unrelated tabular context.

\section{Conclusion}
We introduced \textbf{TabEmbed}, a unified embedding model that bridges the gap between tabular classification and retrieval.
Supported by our proposed \textbf{TabBench} benchmark, we demonstrated that standard text embeddings struggle with tabular structure and numerical semantics.
TabEmbed addresses these challenges through a unified contrastive learning paradigm, utilizing task-adaptive query generation and hard negative mining to learn discriminative representations.
Our experiments reveal that TabEmbed achieves state-of-the-art performance, with the 0.6B model surpassing significantly larger baselines.
This work establishes a baseline for generalist tabular embeddings, demonstrating that rigorous domain alignment is a more effective path to tabular intelligence than parameter scaling alone.
\section*{Limitations}

Despite the promising results of TabEmbed on the proposed benchmark, there are several limitations to our current study.
First, due to the substantial scale of TabBench (comprising over 300 datasets) and budgetary constraints, we did not include commercial closed-source embedding APIs (e.g., Google \texttt{Gemini Embedding}~\cite{lee2025gemini}) in our evaluation.
While our comparison covers a wide range of state-of-the-art open-source models, a comprehensive benchmarking against these commercial systems remains a direction for future research.
Second, our method relies on serializing tabular data into natural language sequences.
For extremely wide tables with hundreds of columns, the serialized text may exceed the maximum context window of the backbone models, potentially leading to information truncation.
Developing more token-efficient serialization strategies or employing long-context architectures to handle ultra-wide tables is an avenue we plan to explore in future work.

\section*{Acknowledgements}
This work was supported by the National Natural Science Foundation of China (No. 62376178), Jiangsu Key Laboratory of Language Computing (JSLCKeyLab 202500003), and the Project Funded by the Priority Academic Program Development of Jiangsu Higher Education Institutions. This work was also supported by Ant Group Research Intern Program.

\bibliography{custom}

\appendix

\section{Related Work}

\subsection{Embedding Models and Tabular Representation}
Text embedding research has evolved from encoder-based architectures to Large Language Models (LLMs).
Early works adapted BERT~\cite{devlin2019bert} and T5~\cite{raffel2020exploring} via contrastive learning~\cite{reimers2019sentence,gao2021simcse}, with recent models like E5~\cite{wang2022text}, BGE~\cite{xiao2024c}, and GTE~\cite{li2023towards} achieving scalability through multi-stage training.
However, limited capacity for complex schemas prompted a shift toward decoder-only LLMs.
Approaches like SGPT~\cite{muennighoff2022sgpt} and E5-Mistral~\cite{wang2024improving} demonstrated the efficacy of generative backbones, while subsequent innovations~\cite{lee2024nv,qwen3embedding} further optimized bidirectional information flow. 
Despite these advancements, existing generalist text models typically process serialized tables as unstructured text, lacking the specific optimization for structural reasoning and numerical understanding required for precise tabular retrieval.

Parallel to text embeddings, representation learning specifically designed for tabular data has also seen significant progress. Early deep tabular models, such as FT-Transformer~\cite{gorishniy2021revisiting} and TabNet~\cite{arik2021tabnet}, introduced specialized attention mechanisms and feature tokenization to handle heterogeneous columns. To achieve transferability across different tables, models like TransTab~\cite{wang2022transtab} and XTab~\cite{zhu2023xtab} proposed shared tokenizers or cross-table pre-training strategies. 
More recently, LLMs have been directly applied to tabular tasks. Approaches such as TabLLM~\cite{hegselmann2023tabllm}, UniPredict~\cite{wang2023unipredict}, and TaPTaP~\cite{zhang2023generative} serialize tabular rows into natural language prompts to perform zero-shot or few-shot classification via text generation. 
However, these methods are primarily confined to generative or predictive paradigms. They either require task-specific architectures or rely on autoregressive decoding for label prediction, failing to produce the dense, fixed-dimensional vectors necessary for efficient semantic search and retrieval in vector databases. 

\subsection{Benchmarks for Text and Tabular Tasks}
Current evaluation protocols remain bifurcated between unstructured text and supervised tabular classification.
In the text domain, standards like BEIR~\cite{thakur2021beir} and MTEB~\cite{muennighoff2023mteb} have driven the progress of retrieval and embedding models, yet they lack dedicated structured data scenarios involving numerical and categorical constraints.
Conversely, tabular machine learning benchmarks, such as OpenML-CC18~\cite{bischl2017openml} and the Grinsztajn suite~\cite{grinsztajn2022tree}, focus primarily on comparing decision trees against neural networks on fixed intra-dataset classification splits. 

Another related line of evaluation includes Table Question Answering (Table QA) and Semantic Parsing benchmarks, such as WikiTableQuestions~\cite{pasupat2015compositional}, Spider~\cite{yu2018spider}, and OTT-QA~\cite{chen2020open}. Pre-trained models like TAPAS~\cite{herzig2020tapas} and TaBERT~\cite{yin2020tabert} were evaluated on these datasets to measure their joint understanding of text and tables. However, these benchmarks are heavily biased toward answering natural language questions over Wikipedia-style, text-heavy tables or translating text to SQL queries. They do not systematically evaluate a model's ability to map raw, heterogeneous tabular rows into a general-purpose embedding space.

While recent massive corpora~\cite{eggert2023tablib,gardner2024large} have enabled large-scale transfer learning suites like TabZilla~\cite{mcelfresh2023neural} and GTL~\cite{wen2024supervised}, these evaluations still treat tabular data learning strictly as an isolated prediction problem. 
To bridge this critical gap, we introduce TabBench, a comprehensive evaluation suite that simultaneously assesses numerical reasoning, semantic alignment, and linear separability, and propose TabEmbed, a generalist model that unifies these diverse tabular tasks within a single shared vector space.

\begin{table}[!tp] 
\centering
\scalebox{0.8}{
\begin{tabular}{l c c c} 
\toprule
\textbf{Model}& \textbf{Layers} & \textbf{Hidden Dim} & \textbf{Max Context} \\ 
\midrule
TabEmbed-0.6B  & 28 & 1024 & 32K \\ 
TabEmbed-4B  & 36 & 2560 & 32K \\ 
TabEmbed-8B & 36 & 4096 & 32K \\ 
\bottomrule
\end{tabular} 
}
\caption{Architectural specifications of the TabEmbed model family. All variants are initialized from the corresponding Qwen3-Embedding checkpoints and inherit their structural configurations.} 
\label{tab:model_architecture} 
\end{table}

\section{Detailed Implementation}
\label{appendix:implementation_details}

\subsection{Training Configurations}
Our training pipeline is built upon the HuggingFace \texttt{Accelerate} library~\cite{accelerate} and the \texttt{Sentence-Transformers} framework~\cite{reimers2019sentence}.
We fine-tune all backbone models for 2 epochs using the contrastive Multiple Negatives Ranking Loss (MNRL) with a temperature parameter $\tau=0.05$.
The training objective is to maximize the cosine similarity between the query $q$ and the positive document $d^+$, while minimizing the similarity with both in-batch negatives and the mined hard negatives described in Section~\ref{sec:triplets}.

We utilize the AdamW optimizer with a learning rate of $1 \times 10^{-5}$ and a linear learning rate decay schedule, following a warmup period spanning the first 10\% of the total training steps.
To balance computational efficiency with the need to capture long-range tabular dependencies, we set the maximum sequence length to 1024 tokens.
The global batch size is set to 256. For the dataset composition, we sample 500,000 retrieval triplets and 100,000 classification triplets from the processed T4 corpus. To ensure training stability, particularly for the 8B parameter models, we employ BFloat16 (BF16) mixed-precision training.

\subsection{Model Architecture}
\label{appendix:architecture}
TabEmbed is built upon the dense decoder-only architecture of the Qwen3-Embedding family.
We release TabEmbed in three sizes (0.6B, 4B, and 8B) to cater to diverse computational constraints.
While our fine-tuning protocol utilizes a context length of 1,024 tokens to optimize training throughput, the underlying architecture supports distinctively long contexts (up to 32K tokens) and variable embedding dimensions.
The detailed architectural specifications for each model variant are summarized in Table~\ref{tab:model_architecture}.

\begin{table}[!tp]
\centering
\scalebox{0.9}{
\begin{tabular}{lrr}
\toprule
\textbf{Category} & \textbf{Count} & \textbf{\# Samples / Corpus} \\
\midrule
\multicolumn{3}{l}{\textsc{Classification Benchmarks}} \\
\midrule
Grinsztajn & 56 & 521,889 \\
OpenML-CC18 & 66 & 249,939 \\
OpenML-CTR23 & 34 & 210,026 \\
UniPredict & 155 & 386,618 \\
\midrule
\textit{Classification Total} & \textit{311} & \textit{1,368,472} \\
\midrule[0.8pt]
\multicolumn{3}{l}{\textsc{Retrieval Benchmarks}} \\
\midrule
Corpus & --- & 1,394,247 \\
Numeric Queries & 10,000 & --- \\
Categorical Queries & 10,000 & --- \\
Mixed Queries & 10,000 & --- \\
\midrule
\textit{Retrieval Total} & \textit{30,000} & \textit{1,394,247} \\
\bottomrule
\end{tabular}
}
\caption{Statistics of the Tabular Embedding Benchmark (TabBench). The benchmark aggregates datasets from four diverse high-quality sources for classification and constructs a large-scale corpus for retrieval tasks.}
\label{table:dataset_stats}
\end{table}

\subsection{Evaluation Protocols}
To ensure rigorous and reproducible evaluation, we fix the random seed to 42 across all experiments.
Table~\ref{table:dataset_stats} summarizes the scale and composition of our evaluation benchmarks (TabBench).
The specific evaluation protocols for the two tasks are as follows:

\paragraph{Tabular Classification (Linear Probing)}
We assess the linear separability of the embedding space by training a lightweight classifier on top of fixed representations.
Specifically, for each evaluated embedding model, we extract frozen dense vectors for all samples. 
Then, for each individual dataset, we train an independent Logistic Regression classifier using the \texttt{scikit-learn} library solely on that dataset's training split embeddings, and evaluate its performance on the corresponding test split. 
It is important to note that there is no shared classifier across datasets; every embedding model is equipped with its own optimized probe per dataset to fairly evaluate its representation quality.
The classifiers are consistently configured with a maximum of 1,000 iterations (\texttt{max\_iter=1000}) and a fixed random seed (\texttt{random\_state=42}) to ensure convergence and strict reproducibility.
We report \textbf{Accuracy} and \textbf{Macro-F1 Score} to account for potential class imbalances in the source datasets.

\paragraph{Tabular Retrieval (Dense Retrieval)}
For the retrieval task, we utilize the \texttt{Faiss} library for efficient vector similarity search.
We employ an exact search strategy using the \texttt{IndexFlatIP} index (Inner Product), which corresponds to Cosine Similarity as all embeddings are $L_2$-normalized prior to indexing.
For each query, we retrieve the top-$k$ most similar documents from the corpus.
Performance is measured using \textbf{MRR@10} (Mean Reciprocal Rank) and \textbf{nDCG@10} (Normalized Discounted Cumulative Gain), which evaluate the ranking quality of the relevant ground-truth rows.
While our main results report metrics at $k=10$, we also compute Recall and Precision at various cutoffs ($k \in \{1, 5, 10, 20, 50, 100\}$) for comprehensive analysis.

\subsection{Heterogeneous Data Serialization Details}
\label{appendix:heterogeneous_data}

Real-world tabular datasets often comprise a wide variety of data types, which traditional schema-bound models (such as tree-based models) handle via specialized encoding layers. To handle such heterogeneous tabular fields within a single generalist architecture, our serialization pipeline unifies all data types into a standardized natural language format prior to string concatenation. The detailed step-by-step procedure is outlined in Algorithm~\ref{alg:data_serialization}.

Specifically, we apply the following pre-processing rules to obtain the string representation $\tilde{v}_j$ for each cell value $v_j$:
\begin{itemize}[leftmargin=.15in]
    \item \textbf{Numeric Data:} Continuous and discrete numerical values (including binary 0/1 integers) are rounded to a specified precision (default two decimal places) to preserve magnitude without excessive tokenization overhead. Integers are preserved without trailing decimals.
    \item \textbf{Categorical and Textual Data:} These fields are converted directly to their literal string representations, stripping trailing periods to avoid punctuation conflicts in the serialization template.
    \item \textbf{Temporal Data:} Dates and timestamps are standardized into the ISO 8601 format (e.g., \textit{``2025-03-01T00:00:00''}) to provide a consistent temporal syntax.
    \item \textbf{Binary Data:} Raw binary streams, where applicable, are decoded into text via UTF-8 or Latin-1 encodings.
\end{itemize}

\begin{figure*}[htbp]
    \centering
    \includegraphics[width=\textwidth]{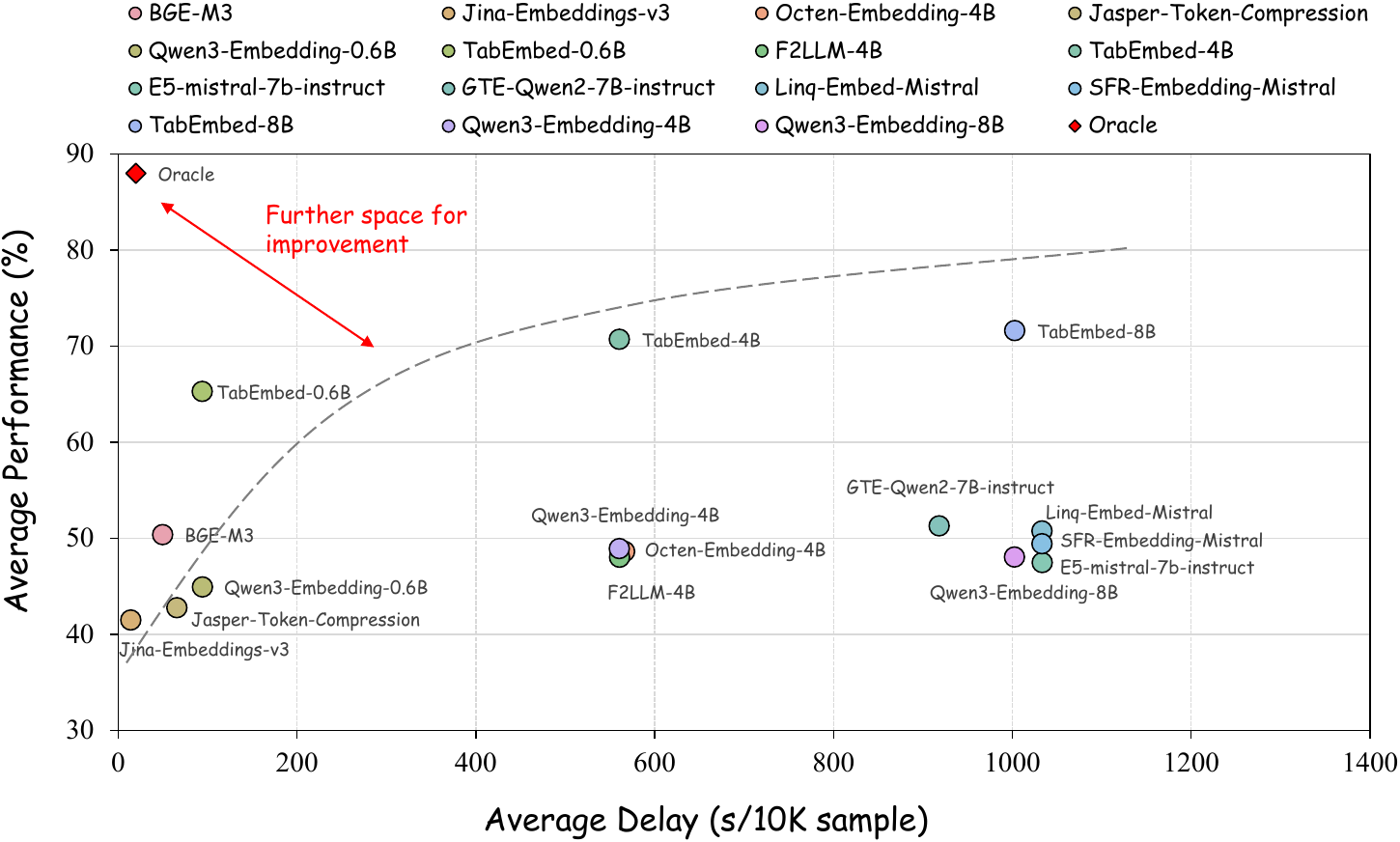}
    \caption{Performance vs. Latency trade-off. The Y-axis represents the overall average performance (computed as the macro-average of the four evaluation metrics) on TabBench, while the X-axis denotes the average inference delay (seconds per 10,000 samples).}
    \label{fig:efficiency_scatter}
\end{figure*}

\begin{algorithm}[!tp]
\caption{Heterogeneous Tabular Data Serialization}
\label{alg:data_serialization}
\textbf{Input:} A tabular row $\mathbf{x} = \{(h_1, v_1), (h_2, v_2), \dots, (h_C, v_C)\}$, numerical precision $\pi$ (e.g., $\pi=2$). \\
\textbf{Output:} Serialized natural language sequence $\mathcal{S}(\mathbf{x})$.
\begin{algorithmic}[1]
\STATE Initialize an empty sequence list $\mathcal{L} \leftarrow \emptyset$.
\FOR{each feature-value pair $(h_j, v_j) \in \mathbf{x}$}
    \IF{$v_j$ is Null \OR NaN}
        \STATE $\tilde{v}_j \leftarrow \text{``unknown''}$
    \ELSIF{$v_j$ is Numeric}
        \STATE $\tilde{v}_j \leftarrow \text{Round}(v_j, \pi)$
        \IF{$\tilde{v}_j$ has no fractional part}
            \STATE $\tilde{v}_j \leftarrow \text{Integer}(\tilde{v}_j)$
        \ENDIF
    \ELSIF{$v_j$ is Date \OR Timestamp}
        \STATE $\tilde{v}_j \leftarrow \text{ISO8601Format}(v_j)$
    \ELSIF{$v_j$ is Binary (Bytes)}
        \STATE $\tilde{v}_j \leftarrow \text{DecodeUTF8}(v_j)$ with fallback to Latin-1
    \ELSE
        \STATE $\tilde{v}_j \leftarrow \text{String}(v_j)$
        \STATE Strip leading/trailing whitespaces and trailing periods from $\tilde{v}_j$.
    \ENDIF
    \STATE $\mathcal{L} \leftarrow \mathcal{L} \cup \{\text{``The } h_j \text{ is } \tilde{v}_j\text{.''}\}$
\ENDFOR
\STATE $\mathcal{S}(\mathbf{x}) \leftarrow \text{Join}(\mathcal{L}, \text{delimiter}=\text{`` ''})$
\STATE \textbf{return} $\mathcal{S}(\mathbf{x})$
\end{algorithmic}
\end{algorithm}

By converting these diverse fields into a unified natural language context, TabEmbed leverages the pre-trained semantic knowledge of the LLM backbone to comprehend heterogeneous data simultaneously, effectively eliminating the need for modality-specific feature engineering.

\subsection{Hardware and Infrastructure}
All experiments are conducted on a high-performance computing cluster equipped with 16 PPU-810E accelerators, each possessing 96GB of high-bandwidth memory.
To efficiently fine-tune the large-scale models (up to 8 billion parameters), we implement a composite optimization strategy.
This includes \textbf{DeepSpeed ZeRO Stage 2} for optimizer state partitioning and \textbf{Gradient Checkpointing} to reduce memory fragmentation.
The multi-GPU training is orchestrated via \textbf{Distributed Data Parallelism}, ensuring linear scaling of the effective batch size.

\begin{figure*}[!tp]
	\centering
	\includegraphics[width=345px]{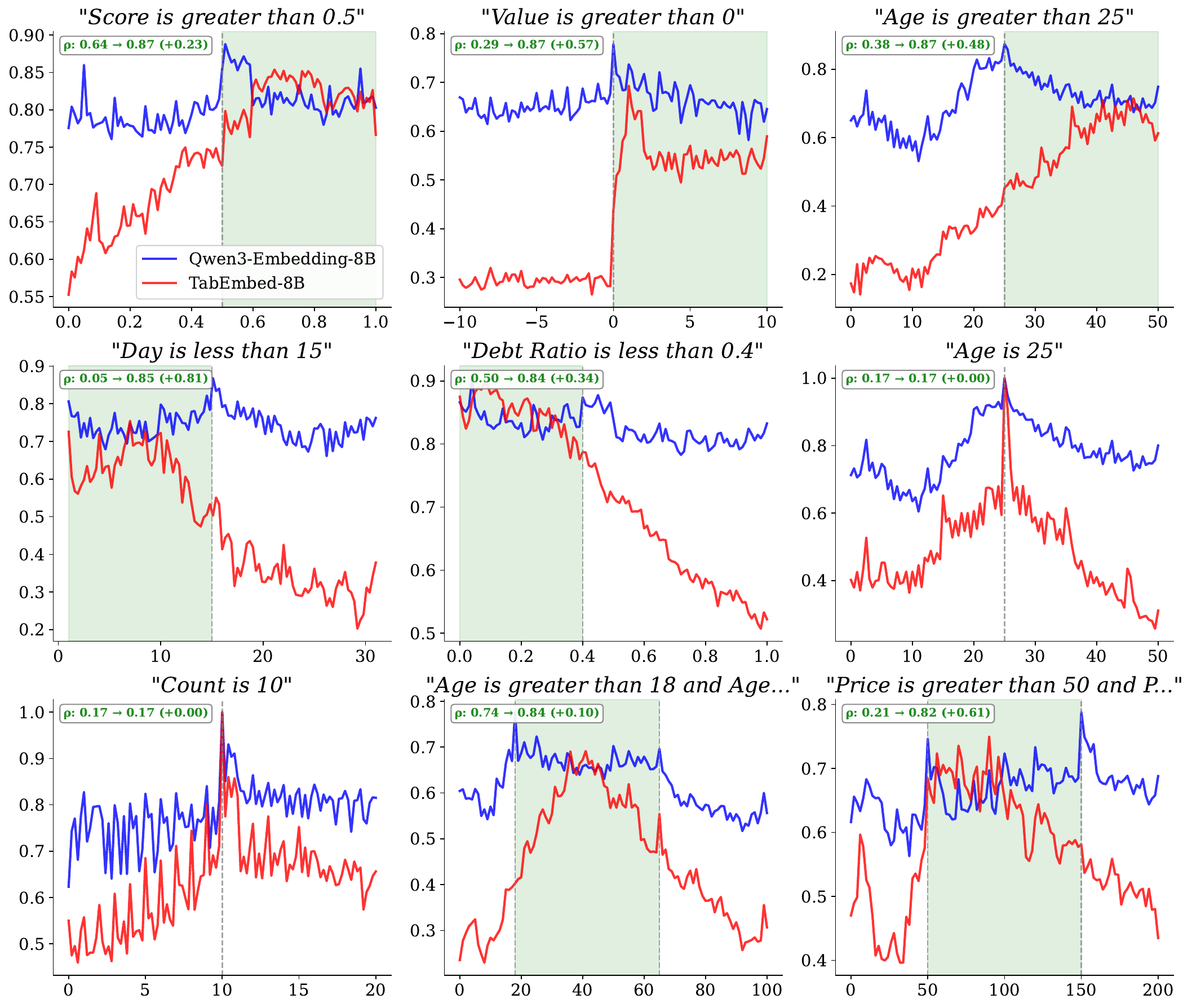}
	\caption{\label{figure:numeric_curves_3x3}  Similarity curves for 9 representative numerical reasoning tasks. \textbf{X-axis}: candidate value; \textbf{Y-axis}: cosine similarity with the query. \textbf{Green shaded regions} indicate valid ranges where conditions are satisfied. \textbf{Blue lines}: baseline Qwen3-Embedding-8B; \textbf{Red lines}: TabEmbed-8B. Spearman correlation ($\rho$) improvements are annotated in each subplot.}
\end{figure*}

\section{Inference Efficiency Analysis}
\label{appendix:efficiency}

To assess the practical viability of TabEmbed for real-world deployment, particularly in resource-constrained environments, we conduct a comprehensive analysis of the trade-off between model performance and inference latency. Figure~\ref{fig:efficiency_scatter} illustrates the relationship between the aggregate performance on TabBench (Y-axis) and the computational cost (X-axis) across different model scales.

We measured inference latency using a standardized benchmarking protocol on a single PPU-810E accelerator. To simulate realistic input distributions comparable to those found in TabBench, we constructed a synthetic dataset comprising serialized tabular rows with lengths varying uniformly between 50 and 200 words. All models were evaluated under identical conditions: a batch size of 64 and a maximum sequence length of 1024 tokens. To ensure statistical stability, we performed a warm-up phase followed by three independent experimental runs, reporting the average latency normalized per 10,000 samples.

The results reveal distinct performance-efficiency clusters corresponding to parameter scales. In the low-latency regime, standard text embedding models such as Jina-Embeddings-v3 and Qwen3-Embedding-0.6B offer high throughput but demonstrate limited capability in capturing tabular semantics, with performance scores hovering around 45\%. \textbf{TabEmbed-0.6B} significantly disrupts this trend, achieving a performance score of 65.27\% while maintaining a highly efficient latency profile ($\approx$ 94 seconds per 10k samples). This indicates that domain-specific contrastive learning can unlock tabular reasoning capabilities in lightweight architectures without incurring additional inference costs.

In the high-capacity regime (4B and 8B parameters), TabEmbed continues to push the performance boundary, reaching up to 71.62\% with the 8B variant. However, this performance gain comes with a considerable increase in computational cost, with latency exceeding 1,000 seconds per 10k samples. TabEmbed-4B offers a compelling middle ground, delivering near-peak performance at approximately half the inference cost of the 8B model. The plot also includes a theoretical "Oracle" point, highlighting the gap that remains between current state-of-the-art models and an ideal system with minimal delay and maximum accuracy. This suggests that future research directions should focus on knowledge distillation or quantization techniques to retain the structural reasoning capabilities of TabEmbed-8B within the latency budget of smaller models.

\begin{figure*}[!tp]
	\centering
	\includegraphics[width=0.96\textwidth]{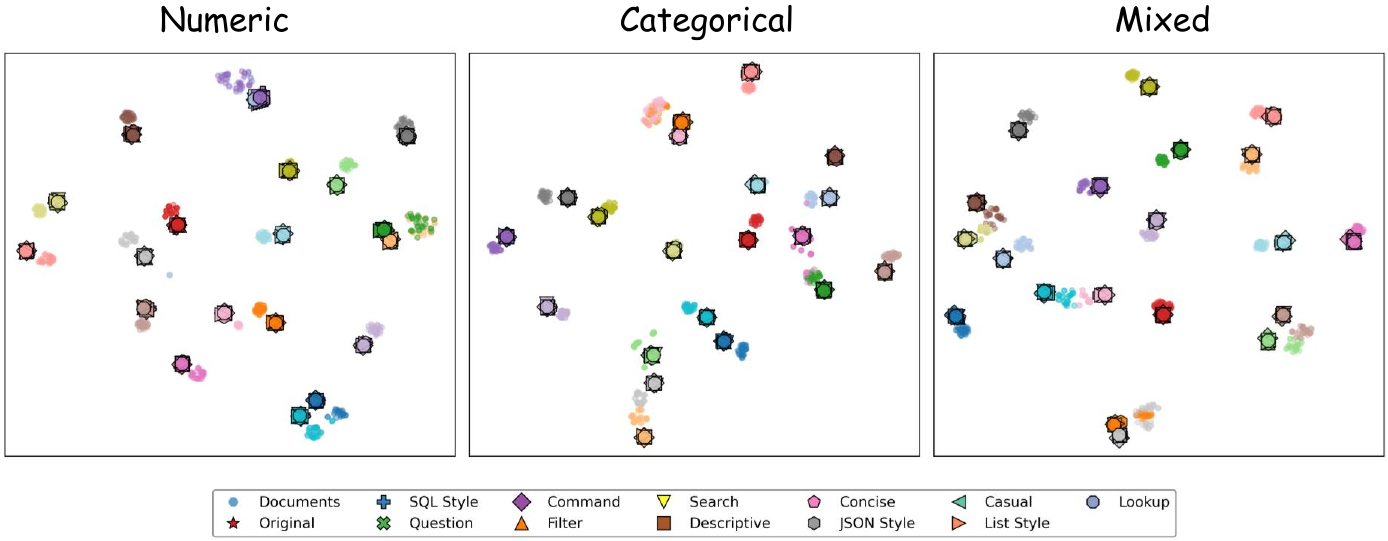}
	\caption{\label{figure:template_robustness}  t-SNE visualization of query template robustness across Numeric, Categorical, and Mixed tasks. For each semantic intent, we generate 12 distinct query variations (e.g., SQL-style, JSON-style, Casual) as defined in Table~\ref{table:query_templates}. The visualization demonstrates that despite significant syntactic differences, all query variations (represented by distinct markers) cluster tightly around the same relevant documents (colored circles), indicating that TabEmbed learns a syntax-agnostic representation of tabular constraints.}
\end{figure*}

\section{Numeric Sensitivity Curves}
\label{appendix:numeric_curves}

To provide a granular view of the numerical reasoning capabilities discussed in Section~\ref{sec:numerical_sensitivity}, Figure~\ref{figure:numeric_curves_3x3} displays the detailed cosine similarity trajectories for nine representative test cases.
For each test case, we define a query $q$ containing a specific numerical constraint (e.g., \textit{``Age is greater than 25''}) and generate a sequence of 101 candidate documents $d(x)$ with values linearly spaced across a relevant range.
We then compute the cosine similarity score regarding the logical truth value: ideal embeddings should yield high similarity only when the condition is met (indicated by the green shaded regions).

As illustrated by the blue lines in Figure~\ref{figure:numeric_curves_3x3}, the baseline Qwen3-Embedding-8B typically exhibits random fluctuations or weak correlations. For instance, in \textit{``Score is greater than 0.5''}, the baseline's similarity scores remain relatively flat or erratic regardless of $x$, confirming that standard text embeddings treat numbers primarily as independent tokens without inherent ordinal semantics.

In contrast, TabEmbed-8B (red lines) demonstrates distinct, logic-aware behaviors tailored to the specific operator types:
\begin{itemize}[leftmargin=*]
    \item \textbf{Inequalities ($>, <$):} The model approximates a step function with sharp transitions at the decision boundary. For example, in \textit{``Age is greater than 25''}, the similarity rises abruptly as $x$ approaches the threshold and sustains a high plateau within the valid range, whereas for \textit{``Day is less than 15''}, it drops significantly once the threshold is exceeded.
    \item \textbf{Equalities ($=$):} For exact matching tasks like \textit{``Age is 25''} or \textit{``Count is 10''}, TabEmbed produces a sharp peak centered exactly at the target value, mimicking a Dirac delta function to distinguish the target from numerically adjacent distractors.
    \item \textbf{Composite Ranges (Between):} For queries involving logical conjunctions (e.g., \textit{``Age is greater than 18...''}), the model accurately delineates the intersection interval, maintaining high similarity only where both conditions hold true.
\end{itemize}

The substantial improvements in Spearman correlation ($\rho$) annotated in each subplot (e.g., $0.64 \to 0.87$) quantitatively verify that TabEmbed has successfully aligned its embedding space with the underlying mathematical logic.

\begin{table*}[!tp]
\centering
\renewcommand{\arraystretch}{1.1}
\begin{tabular}{@{}llp{11.5cm}@{}}
\toprule
\textbf{ID} & \textbf{Template Name} & \textbf{Example Query} \\ 
\midrule
T1 & Original & Find records where age is 25 and salary greater than 50000 \\
T2 & SQL Style & SELECT * FROM table WHERE age = 25 AND salary > 50000 \\
T3 & Question & Which records have age equal to 25 and salary greater than 50000? \\
T4 & Command & Get all entries with age of 25 and salary above 50000 \\
T5 & Filter & Filter: age==25, salary>50000 \\
T6 & Search & Search for records: age:25 salary:>50000 \\
T7 & Descriptive & I need data where the age is 25 and the salary is more than 50000 \\
T8 & Concise & age == 25 | salary > 50000 \\
T9 & JSON Style & \{"age": 25, "salary": \{"\$gt": 50000\}\} \\
T10 & Casual & Show me rows that have age as 25 and salary over 50000 \\
T11 & List Style & Conditions: 1. age equals 25; 2. salary > 50000 \\
T12 & Lookup & Look up records matching: age=25, salary>50000 \\
\bottomrule
\end{tabular}
\caption{Query template variations used to evaluate semantic robustness of embedding models. All templates express the same underlying constraints but differ in linguistic style and format.}
\label{table:query_templates}
\end{table*}

\section{Robustness to Query Template Variations}
\label{appendix:template_robustness}

A critical requirement for a generalist tabular embedding model is the ability to understand the underlying user intent regardless of the input format. Users may express the same retrieval constraint through diverse modalities, ranging from formal syntaxes (e.g., SQL, JSON) to unstructured natural language (e.g., questions, commands).
To evaluate whether TabEmbed has learned a \textbf{syntax-agnostic representation} of tabular constraints, we conducted a qualitative visualization experiment.
We selected representative samples from the Numeric, Categorical, and Mixed retrieval tasks. For each sample, we generated 12 distinct variations using the templates listed in Table~\ref{table:query_templates}, effectively creating a set of semantic equivalence classes where queries differ in surface form but share identical logical constraints.

Figure~\ref{figure:template_robustness} presents the t-SNE projection of these embeddings.
The visualization reveals a striking geometric pattern: for every semantic intent, the diverse query variations (represented by distinct markers such as stars, crosses, and triangles) form tight, cohesive clusters surrounding their corresponding ground-truth documents (colored circles).
Notably, this alignment persists across extreme syntactic disparities.
For instance, highly structured formats like \textbf{JSON Style} (T9: {"age": 25...}) and \textbf{SQL Style} (T2: SELECT * FROM...) are mapped to the immediate vicinity of unstructured natural language queries like \textbf{Casual} (T10) and \textbf{Question} (T3).
This observation confirms that TabEmbed does not merely rely on keyword matching or rigid template overfitting.
Instead, it has successfully learned to extract the \textbf{invariant logical semantics} (e.g., numerical magnitude and equality constraints) from the input, projecting semantically equivalent queries to the same point on the manifold regardless of their linguistic style.
This capability ensures that TabEmbed can generalize to diverse real-world search scenarios where user querying habits may vary significantly.

\begin{figure}[!tp]
	\centering
	\includegraphics[width=\columnwidth]{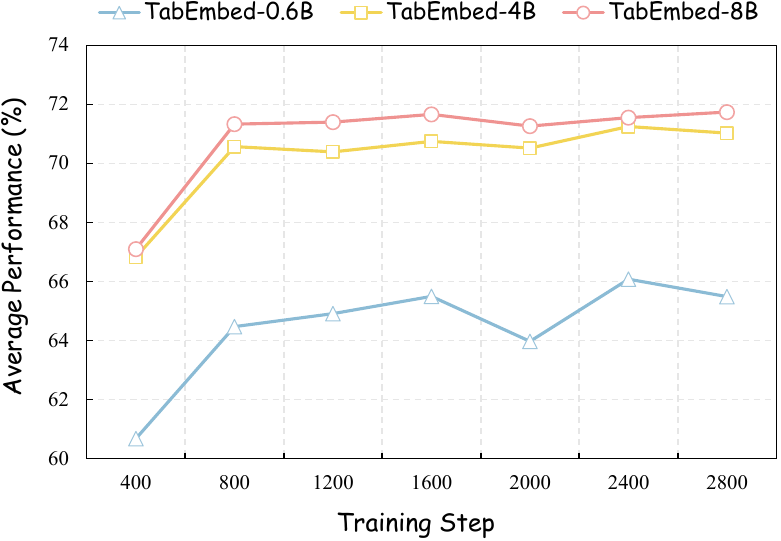}
	\caption{\label{figure:training_steps}  The impact of training steps on the average performance (the macro-average of the four evaluation metrics) across TabBench. We track the evaluation metrics of TabEmbed at different parameter scales (0.6B, 4B, and 8B) from 400 to 2800 training steps.}
\end{figure}

\section{Training Convergence Analysis}
\label{appendix:convergence}
To determine the optimal training duration and investigate the convergence behavior of our unified contrastive learning paradigm, we monitor the average performance of TabEmbed across three parameter scales (0.6B, 4B, and 8B) at regular intervals, ranging from 400 to 2800 training steps.

As illustrated in Figure~\ref{figure:training_steps}, we observe several key findings regarding training efficiency and model scaling.
First, Rapid Convergence. All models exhibit a steep performance trajectory in the initial phase. Notably, a significant portion of the performance gain is realized within the first 800 steps.
For instance, the 4B model improves from roughly 67\% to over 70\% in this short period.
The performance curves generally plateau after approximately 1600 steps, suggesting that our framework is highly data-efficient and does not require excessively prolonged training to learn robust tabular representations.
Second, Impact of Model Scale. Consistent with neural scaling laws, larger models consistently achieve higher performance ceilings.
TabEmbed-8B (pink line) maintains a clear superiority over the 4B and 0.6B variants throughout the training process.
Furthermore, model scale correlates positively with training stability. The 4B and 8B models display smooth and monotonic improvements, whereas the 0.6B model (blue line) exhibits noticeable volatility, particularly around step 2000.
This suggests that larger foundation models possess a more robust latent space, making them less susceptible to batch noise during contrastive optimization.

\begin{algorithm*}[htbp]
\small
\caption{Dynamic Target Identification and Discretization}
\label{alg:target_selection}
\textbf{Input:} A tabular dataset $\mathbf{T}$ with columns $\mathcal{C} = \{c_1, c_2, \dots, c_M\}$, categorical sampling probability $p$ (e.g., $p=0.5$). \\
\textbf{Output:} Selected target column $y$ and its processed discrete labels.
\begin{algorithmic}[1]
\STATE Initialize valid candidate set $\mathcal{V} \leftarrow \emptyset$.
\FOR{each column $c \in \mathcal{C}$}
    \STATE $U_c \leftarrow \text{UniqueValues}(c)$
    \STATE \textit{\% Check against all rejection criteria}
    \IF{$\text{Name}(c)$ contains ``Unnamed:'' \OR $\text{Type}(c)$ is Date/Time}
        \STATE \textbf{continue}
    \ELSIF{$|U_c| < 2$ \OR $|U_c| > 50$ \OR $\max_{v \in U_c} \text{Length}(v) > 256$}
        \STATE \textbf{continue}
    \ELSIF{$|U_c| == \text{Rows}(\mathbf{T})$ \AND $c$ is not strictly numeric}
        \STATE \textbf{continue}
    \ELSE
        \STATE $\mathcal{V} \leftarrow \mathcal{V} \cup \{c\}$
    \ENDIF
\ENDFOR
\STATE Partition $\mathcal{V}$ into numerical candidates $\mathcal{V}_{\text{num}}$ and categorical candidates $\mathcal{V}_{\text{cat}}$.
\IF{$\mathcal{V}_{\text{num}} \neq \emptyset$ \AND $\mathcal{V}_{\text{cat}} \neq \emptyset$}
    \STATE Sample $y$ uniformly from $\mathcal{V}_{\text{cat}}$ with probability $p$, otherwise from $\mathcal{V}_{\text{num}}$.
\ELSE
    \STATE Sample $y$ uniformly from $\mathcal{V}$.
\ENDIF
\IF{$y \in \mathcal{V}_{\text{num}}$}
    \STATE $\mathcal{Q} \leftarrow \text{Calculate 4-quantiles for } y$
    \STATE Discretize continuous values in $y$ into 4 buckets based on $\mathcal{Q}$.
    \STATE Convert each bucket into natural language descriptors (e.g., ``less than $q_1$'').
\ENDIF
\STATE \textbf{return} $y$
\end{algorithmic}
\end{algorithm*}

\section{Target Column Selection Criteria}
\label{appendix:target_selection}

To transform the unannotated tables from the T4 corpus into high-quality supervised classification tasks, we employ a dynamic target identification pipeline. The complete procedure is summarized in Algorithm~\ref{alg:target_selection}.

For a given table $\mathbf{T}$, we first consider all columns as potential target candidates and then apply a rigorous filtering protocol to exclude non-informative or trivial prediction targets. Specifically, a column $c$ is excluded from the candidate pool if it satisfies any of the following rejection criteria:

\begin{itemize}[leftmargin=.15in]
    \item \textbf{Low Informativeness:} The column contains only a single unique value (constant columns), providing no discriminative signal.
    \item \textbf{High Cardinality / Identifiers:} The column possesses a unique value for every row (e.g., UIDs, Row IDs), or the number of unique classes exceeds 50. Such columns typically lead to trivial memorization rather than semantic generalization.
    \item \textbf{Data Type Constraints:} The column is identified as a date/timestamp, or contains textual values exceeding 256 characters, which are unsuitable for standard classification objectives.
    \item \textbf{Column Name Constraint:} The column name contains ``Unnamed:'' (pandas' default marker for unnamed columns).
\end{itemize}

It is important to note that columns failing these criteria are only excluded from being selected as the \textit{prediction target} $y$. They remain part of the input feature set $\mathbf{x}_{-y}$ to provide context, unless they are removed by standard feature selection processes.

Once the set of valid candidate columns is established, we select a single target $y$ for each training instance. To balance the distribution of task types, we employ a weighted sampling strategy. Based on our qualitative observation that categorical columns often yield higher-quality decision boundaries than arbitrary numerical regression targets, we prioritize classification tasks. Specifically, if both continuous and categorical candidates are present, we sample a categorical target with probability $p=0.5$ and a continuous target with probability $1-p=0.5$.

In cases where a continuous column is selected as the target, we transform the regression problem into a classification problem via dynamic discretization. We divide the continuous range into quantile-based bins (defaulting to 4 buckets) to ensure class balance. The resulting targets are serialized into natural language class descriptors, such as \textit{``less than 15.5''}, \textit{``between 15.5 and 40.2''}, or \textit{``greater than 40.2''}. This unified serialization allows TabEmbed to handle both original categorical labels and discretized numerical bins within the same semantic embedding space.

\section{Baselines}
\label{appendix:baselines}

We compare TabEmbed against a diverse set of state-of-the-art generalist text embedding models, ranging from lightweight encoders to large-scale LLM-based embeddings.

\paragraph{0.6B Parameter Scale}
\begin{itemize}[leftmargin=.15in]
    \item \textbf{Jina-Embeddings-v3}~\cite{sturua2024jina}: A multilingual, multi-task embedding model based on the Jina-XLM-RoBERTa architecture. It incorporates Rotary Position Embeddings (RoPE) to support extended context windows up to 8192 tokens. A key feature of this model is the integration of five task-specific LoRA adapters, allowing for efficient generation of embeddings tailored to specific downstream applications.
    
    \item \textbf{Jasper-Token-Compression}~\cite{zhang2025jaspertokencompression600mtechnicalreport}: A 600M parameter model from the Jasper series that introduces dynamic text token compression, inspired by DeepSeek-OCR strategies. By combining vector distillation with contrastive learning, it achieves high performance while compressing textual information by approximately 10x, offering a unique approach to efficient representation.
    
    \item \textbf{Qwen3-Embedding-0.6B}~\cite{qwen3embedding}: The lightweight variant of the latest proprietary embedding series from the Qwen family. Building upon the dense Qwen3 foundational architecture, it inherits strong multilingual capabilities and reasoning skills, optimized specifically for retrieval and ranking tasks.
\end{itemize}

\paragraph{4B Parameter Scale}
\begin{itemize}[leftmargin=.15in]
    \item \textbf{F2LLM-4B}~\cite{zhang2025f2llm}: Standing for ``Foundation to Feature Large Language Models,'' F2LLM is fine-tuned on a curated corpus of 6 million high-quality query-document pairs sourced exclusively from open-source datasets. It employs a single-stage training process with homogeneous macro batches, eschewing complex multi-stage pipelines while covering diverse retrieval and clustering tasks.

    \item \textbf{Octen-Embedding-4B}~\cite{octen2025rteb}: Built upon the Qwen3 foundation, this model is specifically optimized for complex, real-world industry retrieval scenarios. Its training pipeline leverages large-scale domain-specific synthetic data across legal, finance, healthcare, and code domains. By employing parameter-efficient LoRA fine-tuning, cross-device negative sharing, and multi-domain model fusion, it achieves a strong balance between retrieval performance and computational efficiency while supporting ultra-long contexts of up to 32,768 tokens.
    
    \item \textbf{Qwen3-Embedding-4B}~\cite{qwen3embedding}: A mid-sized model in the Qwen3 embedding series. It balances computational efficiency with the advanced long-text understanding capabilities of the Qwen3 foundation, serving as a strong baseline for mid-scale generalist text embeddings.
\end{itemize}

\paragraph{7B-8B Parameter Scale}
\begin{itemize}[leftmargin=.15in]
    \item \textbf{SFR-Embedding-Mistral}~\cite{SFRAIResearch2024}: Developed by Salesforce Research, this model is initialized from E5-Mistral-7b-instruct and Mistral-7B-v0.1. It represents a robust baseline for instruction-tuned embeddings derived from decoder-only architectures.
    
    \item \textbf{Linq-Embed-Mistral}~\cite{LinqAIResearch2024}: Also built upon the E5-Mistral and Mistral-7B foundations, Linq-Embed-Mistral focuses on enhancing retrieval performance through advanced data refinement. Its training pipeline emphasizes sophisticated data crafting, rigorous filtering, and hard-negative mining guided by teacher models to improve the quality of synthetic training triplets.
    
    \item \textbf{GTE-Qwen2-7B-Instruct}~\cite{li2023towards}: The latest addition to the General Text Embedding (GTE) family, built on the Qwen2-7B LLM. It leverages the same training data and strategies as its predecessor (GTE-Qwen1.5) but benefits from the architectural upgrades of the Qwen2 base model. It is a leading performer on the MTEB benchmark, particularly in multilingual evaluation scenarios.
    
    \item \textbf{Qwen3-Embedding-8B}~\cite{qwen3embedding}: The largest model in our comparison suite and the direct backbone for TabEmbed-8B. It represents the state-of-the-art in the Qwen family for dense retrieval, bitext mining, and classification, providing a rigorous baseline to measure the impact of our domain-specific contrastive learning paradigm.
\end{itemize}

\section{Applications}
\label{appendix:applications}

The unified representation capability of TabEmbed and the comprehensive evaluation framework of TabBench open up several promising avenues for real-world applications. In particular, it complements existing tabular reasoning systems by providing a foundational, schema-agnostic semantic layer.

\subsection{Foundational Retrieval Layer for Agentic RAG Systems}
Recent advancements in agentic RAG and Text-to-SQL pipelines have demonstrated strong capabilities in performing complex tabular reasoning, such as aggregations and table joins, when the database schema is well-defined. 
However, Text-to-SQL approaches face significant challenges in scenarios involving unstructured queries, fuzzy matching, or instances where the user is unaware of the underlying schema.
Instead of replacing these reasoning agents, TabEmbed serves as a crucial complementary \textbf{Foundational Retrieval Layer}. By mapping natural language constraints (e.g., \textit{``High-value users from the tech sector''}) and structured rows into a shared vector space, it enables databases to perform millisecond-level row retrieval via vector similarity search. This effectively acts as a high-efficiency, cost-effective filter that retrieves relevant context before passing it to downstream LLM agents for heavier logical reasoning.

\subsection{Enterprise Data Discovery and Data Lakes}
Large enterprises often maintain massive data lakes containing thousands of unorganized spreadsheets and CSV files (the "dark data" problem). 
In such massive-scale scenarios, running an LLM agent to analyze schemas and generate SQL for every query is computationally prohibitive and incurs unacceptable latency.
TabEmbed overcomes this bottleneck by enabling \textbf{Offline Indexing and Semantic Data Discovery}. By embedding sample rows or summarized schemas from thousands of tables into a unified vector index, users can search for datasets using vague intent queries (e.g., \textit{``I need sales data regarding Q3 revenue in Southeast Asia''}). Unlike strict keyword-based search or exact SQL matching, TabEmbed understands the numerical and categorical semantics within the table content, efficiently locating relevant tables even if the column headers do not explicitly match the query keywords.

\subsection{Zero-Shot and Cold-Start Tabular Prediction}
In many dynamic industrial applications (e.g., fraud detection in new markets, user churn prediction for new products), historical training data is scarce or unavailable. Traditional supervised models (like XGBoost) are rendered ineffective in these \textbf{Cold-Start Scenarios} because they require task-specific retraining on fixed schemas.
As demonstrated by our classification experiments, TabEmbed possesses strong zero-shot transfer capabilities. It can function as a generic feature extractor or a nearest-neighbor classifier right out of the box. Practitioners can simply convert a handful of labeled examples (a support set) into vectors and classify new incoming rows based on embedding similarity, enabling immediate predictive capabilities without the need for time-consuming feature engineering or model training.

\subsection{Cross-Schema Entity Resolution and Data Integration}
A pervasive challenge in database management is \textbf{Entity Resolution} (or Record Linkage), identifying rows across different databases that refer to the same real-world entity despite having different schemas, missing values, or inconsistent naming conventions (e.g., merging a ``Client'' table with a ``Customer'' table after a corporate acquisition). 
Traditional methods rely heavily on manual schema matching and hand-crafted string similarity rules. 
Because TabEmbed serializes diverse tabular fields into a unified natural language context, it inherently learns a schema-agnostic representation. Two rows describing the same entity with different column names or formatting will be projected into close proximity within the embedding space. Consequently, enterprise data integration can be elegantly reformulated as a cross-database vector similarity search, bypassing the arduous process of manual schema alignment.

\subsection{Semantic Anomaly Detection and Data Cleaning}
Real-world tabular data is notoriously noisy, often containing logical contradictions (e.g., a "Status: Active" subscription with a "Termination Date" in the past) or numerical errors. Rule-based data cleaning requires domain experts to anticipate and hardcode every possible error type.
TabEmbed offers a robust alternative for out-of-the-box \textbf{Semantic Anomaly Detection}. By projecting all rows of a table into the learned embedding space, standard density-based anomaly detection algorithms (such as Isolation Forest or Local Outlier Factor) can be directly applied to the dense vectors. Since TabEmbed is pre-trained to understand structural logic and numerical magnitude, rows containing semantic contradictions or extreme outliers will naturally isolate themselves in the manifold. This provides a zero-shot, automated data cleaning mechanism that does not rely on predefined schemas or rules.

\end{document}